\documentclass{ifacconf}

\usepackage{graphicx}      
\usepackage{natbib}

\usepackage{mathrsfs}
\usepackage{color}
\usepackage{amsfonts,amssymb}
\usepackage{amsfonts}
\usepackage{subfigure}
\usepackage{booktabs}

\usepackage{amsmath}
\usepackage{enumerate}
\usepackage[ruled,vlined,algo2e]{algorithm2e}
\SetKwRepeat{Do}{do}{while}%
\usepackage{bm}
\usepackage{makecell}

\usepackage{mathtools}
\usepackage{bbm}
\usepackage{dsfont}
\usepackage{pifont}

\usepackage{enumitem}
\usepackage{balance}

\newtheorem{remark}{Remark}

\newtheorem{lemma}{Lemma}

\newtheorem{problem}{Problem}

\begin{document}

\begin{frontmatter}
\title{Control Input Inference of Mobile Agents under Unknown Objective}

\thanks[footnoteinfo]{The work of X. Duan is sponsored by Shanghai Pujiang Program under grant 22PJ1404900.}

\author{Chendi Qu, Jianping He, Xiaoming Duan, Shukun Wu
} 

\address[First]{The Dept. of Automation, Shanghai Jiao Tong University, and Key Laboratory of System Control and Information Processing, Ministry of Education of China, Shanghai, China. \\(e-mail: qucd21, jphe, xduan, wsk798@sjtu.edu.cn)}

\begin{abstract}
Trajectory and control secrecy is an important issue in robotics security.
This paper proposes a novel algorithm for the control input inference of a mobile agent without knowing its control objective. Specifically, the algorithm first estimates the target state by applying external perturbations. Then we identify the objective function based on the inverse optimal control, providing the well-posedness proof and the identifiability analysis. Next, we obtain the optimal estimate of the control horizon using binary search. Finally, the agent's control optimization problem is reconstructed and solved to predict its input. Simulation illustrates the efficiency and the performance of the algorithm.\vspace{-8pt}
\end{abstract}
\begin{keyword}
Mobile Agents, Control Input Inference, Optimal Control, Trajectory Secrecy
\end{keyword}
\end{frontmatter}

\section{Introduction}
\vspace{-10pt}
As technologies of perception, localization and planning gradually mature, mobile agents have been widely applied and achieved high-profile success \citep{cao2012mobile}. One of the basic application scenarios is to control an agent from one place to another.
Accordingly, preserving the security and the trajectory secrecy of the agents becomes a critical issue \citep{pasqualetti2013attack}. 

Consider a mobile agent driving from its initial position to a target point, whose states are observed and recorded by an external attacker. The attacker wants to learn the motion of the agent and predict the trajectory accurately for subsequent attacks. Some existing methods are data-driven and model-free, such as using polynomial regression \citep{gao2018online} or introducing neural networks for the prediction, including the long short-term memory neural network \citep{altche2017lstm} or graph neural network \citep{mohamed2020social}. However, these require a large amount of observations as well as the model training in the early stage. Another class of algorithms is model-based.
For instance, \cite{schulz2018multiple} use an unscented Kalman filter to predict multi-agent trajectories. \cite{li2020unpredictable} measure the secrecy of the trajectory and proves that uniformly distributed inputs maximize the unpredictability of the system. 
However, these predictive models often assume that the control inputs are known.
Notice that if we can learn the control law of the agent, then we are able to infer the input in any situations and predict the future trajectory accurately, combining with the estimation of the dynamic model and current state through System Identification (SI) \citep{ljung1998system} and data fusion filters \citep{gillijns2007unbiased}, respectively. 

This paper studies the control input inference of the mobile agent, which is a non-trivial problem since the control objective of the agent is unknown. Different from the input inference described in \cite{watson2020stochastic}, which is used to generate optimal control of stochastic nonlinear systems, we try to reconstruct the control optimization problem by inferring the objective of the mobile agent and further infer the inputs.
Suppose we have obtained the dynamic model with SI, in order to reconstruct the optimization problem of the agent, we still need to estimate three things: the target state, the objective function parameters and the control horizon.
The idea to reconstruct the optimization problem given the optimal state trajectories is similar to the inverse optimal control (IOC) \citep{ab2020inverse, yu2021system, molloy2020online}. Yet, our work is more general since IOC is mainly utilized to identify objective function parameters which is a part of the problem reconstruction. 

Motivated by the above discussion, we design a novel algorithm for the control input inference of mobile agents with unknown objective. 
This algorithm can not only help us predict the input and state trajectory of the mobile agent precisely, which serves as the basis for the subsequent attack or interception, but also learn the control law and simulate the motion of the agent. The challenge lies in inferring the unknown control objective, which contains several parts, based on trajectory observations with noises.
The main contributions are summarized as follows:
\begin{itemize}
\item We propose an algorithm on the control input inference of a mobile agent with unknown objectives. Through estimating the target state, objective function parameters and the control horizon, the algorithm reconstructs the control optimization problem and calculates the future inputs of the agent.
\item In order to identify the quadratic objective function, we formulate and solve an inverse optimal control problem given state trajectories in the presence of observation noises. Well-posedness proof and identifiability analysis are also provided. Moreover, a novel binary search method for estimating the agent's control horizon is presented, along with the analysis of the algorithm efficiency.
\end{itemize}

The remainder of the paper is organized as follows. Section \ref{preliminary} describes the problem of interest. Section \ref{target} and \ref{ioc} study the target point estimation and objective function identification. Section \ref{horizon} estimates the control horizon and presents the full algorithm. Simulation results are shown in Section \ref{sim}, followed by the conclusion in Section \ref{conc}. 

\vspace{-8pt}
\section{Preliminaries and Problem Formulation}\label{preliminary}
\vspace{-8pt}
\subsection{Model Description}
\vspace{-8pt}
Consider a mobile agent $R_M$ driving from its initial state to a fixed target point $x_T$. $R_M$ is modeled by a discrete-time linear system
\begin{equation}\label{sys}
\left\{
\begin{array}{ll}
{x}_{k+1} = A {x}_k + B {u}_k,\\
{y}_{k} = {x}_k + \omega_k,
\end{array}
\right.
\end{equation}
where ${x}_k$ is the state vector, ${y}_k$ is the outputs such as agent's position, ${u}_k$ is the control input and $A, B$ are $n \times n$ and $ n \times m$ matrices. 
$\omega_k$ in \eqref{sys} is zero mean observation noise with $\mathbb{E}(\Vert \omega_k \Vert^2) < + \infty$.

Assume that (A, B) is controllable and B has
full column rank. Moreover, $A$ is an invertible matrix since the system matrix of a discrete-time system sampled from a continuous linear system is always invertible \citep{zhang2019inverse}. 

Note that $R_M$ follows the optimal control and the control optimization problem is described as
\vspace{-10pt}
\begin{subequations}\label{eq:op_pro1}
\begin{eqnarray}\nonumber
\mathbf{P}_0: & & \min_{u_{0:N-1}} \,J = \frac{1}{2}\{x_N^T H x_N + \sum_{k = 0}^{N-1}  u_k^T R u_k\}, \\ 
&& \mathrm{s.t.} ~~
\eqref{sys}, x_0 = \bar{x}- x_T,\,k = 0,1,\dots,N-1,\nonumber
\end{eqnarray}
\end{subequations}
where $H, R$ are positive definite matrices and $\bar{x}$ is the initial state. The first term in $J$ reflects the deviation from the target state and the second term represents the energy cost during the process. The solution of $\mathbf{P}_0$ is
\begin{equation}\label{eq:uk}
u_k = -K_k x_k,\,k = 0,1,\dots,N-1,
\end{equation}
where $K_k$ is the control gain matrix related to the system equation and control objectives.
We assume that when $R_M$ receives an external disturbance and deviates from its original trajectory, it will recalculate the optimal control by setting the current state as the initial point.
\begin{remark}
Note that the objective function in $\mathbf{P}_0$ omits the term $x_k^T Q x_k$ since the agent mainly pays attention to reaching the target state and the energy consumption instead of transient states. Moreover, to facilitate the identification, we assume the target state is $0$ in $\mathbf{P}_0$ and perform a coordinate transformation on the initial state $\bar{x}$.
\end{remark}
\vspace{-8pt}
\subsection{Problem Formulation}
\vspace{-8pt}
Suppose there is another external agent $R_A$ observing and recording the output trajectories of $R_M$. 
$R_A$ has exact knowledge of the dynamic function parameters $A, B$, which can be obtained by SI methods, and the quadratic form of the objective function is known as well.
Suppose that $R_A$ obtains $M \geqslant n$ optimal trajectories of $R_M$. Denote the $j$-th trajectory as $y^j = \{y_0^j, y_1^j,\dots,y_{N_j}^j\}$, $j = 1, 2, \dots, M$, where $N_j \geqslant 2n$ is the length of the trajectory. We assume the initial state is free of observation noises, i.e., $y_0^j = x_0^j$. We require that there exist at least $n$ linearly independent final states among the data, which means the matrix $[y_{N_1}^1\, \dots \, y_{N_M}^M]$ has a full row rank.
\begin{remark}\label{rem_tr}
According to our previous assumption, $R_M$ will recalculate the trajectory after receiving disturbances. Therefore, we can leverage this mechanism and actively apply external inputs to collect different trajectories.
\end{remark}

At time $k = l$, $R_A$ has observed $R_M$ for $l+1$ steps and obtained a series of observations 
$y = \{y_0, y_1,\dots, y_l\}$.
Denote the input inference by $\hat{u}_{l|l}$. $R_A$ tries to infer the current control input accurately, which is described by
\begin{equation}
\min \Vert u_l - \hat{u}_{l|l} \Vert^2 = \Vert -K_l x_l - \hat{u}_{l|l}\Vert^2,
\end{equation}
where $u_l$ is the real input at $k=l$.
We can obtain the unbiased estimate of the state $\hat{x}_{l|l}$ with the filter in \cite{gillijns2007unbiased}. Then the main problem is to estimate $K_l$ accurately and $\hat{u}_{l|l} = \hat{K}_{l|l} \hat{x}_{l|l}$.

However, under our assumption, the control objective of $R_A$ is unknown. Therefore, in order to predict the control input at $l$, we need to estimate: i) the target state ${x}_T$; ii) the objective function parameters ${H}, {R}$; iii) the control horizon ${N}$. Then, we can reconstruct the optimization problem of $R_A$. If the above estimates are accurate, we can obtain the optimal inference $\hat{u}_{l|l}^*$ by solving the reconstructed optimization problem. 
Hence, in the following sections, we will estimate the above parameters and then formulate the optimization problem to predict the control inputs.

\vspace{-8pt}
\section{Target Point Estimation}\label{target}
\vspace{-8pt}
In this section, we will estimate the target position of $R_M$.
We choose at least two non-parallel trajectories and calculate the intersection of their extension lines as the target estimation. 

Assume $R_M$'s current position is $p_0^1 \in \mathbb{R}^{2}/\mathbb{R}^{3}$. Observe and record the trajectory for $L_1+1$ steps, which is denoted as $\{p_0^1, p_1^1, \dots, p_{L_1}^1\}$. By Remark \ref{rem_tr}, when we add an external stimulus $u_a$, the state of $R_M$ changes into $p_0^2$. Similarly, record $\{p_0^2, p_1^2, \dots, p_{L_2}^2\}$. It is required that two trajectories are not parallel, i.e., the vectors $p_0^1 - p_{L_1}^1$ and $p_0^2 - p_{L_2}^2$ are linearly independent. 
Then we need to fit the lines where the trajectories lie and solve for their intersection. The solution of the 2D case is presented in \cite[Sec IV.B]{9882307}. Here we will provide the algorithm for the 3D case.

\textbf{Line Fitting:}
Let $p_i^j = (p_{x,i}^j, p_{y,i}^j, p_{z,i}^j)^T$. We have the equation of a straight line in 3D space as
\begin{align}\nonumber
& p_x = \frac{a}{c} (p_z - z_0) + x_0 = m_1 p_z + n_1, \\ \nonumber
& p_y = \frac{b}{c} (p_z - z_0) + y_0 = m_2 p_z + n_2.
\end{align}
Omit the trajectory number $j$ for brevity. 
To minimize the sum of squares of the residuals, we have
\vspace{-3pt}
\begin{align}\nonumber
\min \sum_i^{L_1} (p_{x,i} - m_1 p_{z,i} - n_1)^2
\vspace{-5pt}
\end{align}
for $p_x$. Taking derivative with respect to $m_1$ and $n_1$ and setting the derivatives equal to 0, we obtain
\begin{align}\nonumber
& m_1 = \frac{L_1 \sum_i p_{x,i} p_{z,i} - \sum_i p_{x,i} \sum_i p_{z,i}}{L_1 \sum_i {p_{z,i}}^2 - \sum_i p_{z,i} \cdot \sum_i p_{z,i}},\\ \nonumber
& n_1 = \frac{\sum_i p_{x,i} - m_1 \sum_i p_{z,i}}{L_1}.
\end{align}
The calculation of $m_2,n_2$ is similar, except that $p_{x,i}$ is replaced by $p_{y,i}$.

\textbf{Calculate the ``Intersection":}
Suppose two straight lines $l_1, l_2$ obtained by line fitting have direction vectors $v_1, v_2$, respectively. They are guaranteed not to be parallel, i.e., 
$v_1 \times v_2 \neq 0$.
However, due to the existence of the observation noises, the two fitted lines may not intersect. In this case, when the shortest distance $d_{l_1,l_2}$ between the lines satisfies
\begin{equation}\label{distance}
d_{l_1,l_2} \leqslant \epsilon, \, \epsilon \in \mathbb{R}_+,
\end{equation}
we consider the two lines intersect and take the midpoint of the nearest two points on $l_1,l_2$ as an estimate of the intersection. 

Let $v_3 =\overrightarrow{o_1 o_2}$, where $o_1,o_2$ are points of $l_1,l_2$. Then we obtain
\vspace{-5pt}
\[
d_{l_1,l_2} = \frac{(v_1 \times v_2) \cdot v_3}{\Vert v_1 \times v_2 \Vert}.
\]
If \eqref{distance} holds, then we have
\begin{equation}\nonumber
\left\{
\begin{array}{ll}
l_1(t_1) = o_1 + t_1 \cdot v_1,\\
l_2(t_2) = o_2 + t_2 \cdot v_2,
\end{array}
\right.
\end{equation}
and
\vspace{-3pt}
\[
[l_1(t_1) - l_2(t_2)] \cdot v_1 = 0,\,[l_1(t_1) - l_2(t_2)] \cdot v_2 = 0.
\]
We solve for $t_1$ and $t_2$ as
\begin{align}\nonumber
& t_1 = \frac{(v_1 \cdot v_2) (o_2-o_1) \cdot v_2 - (v_2 \cdot v_2) (o_2-o_1) \cdot v_1}{(v_1 \cdot v_2)^2 - (v_1 \cdot v_1) (v_2 \cdot v_2)}, \\ \nonumber
& t_2 = \frac{v_1 \cdot v_1}{v_1 \cdot v_2} t_1 - \frac{(o_2-o_1) \cdot v_1}{v_1 \cdot v_2}.
\end{align}
Then the intersection is calculated by $\frac{l_1(t_1) + l_2(t_2)}{2} := \hat{x}_T$.

\begin{figure}[t]
\centering
\includegraphics[width=0.25\textwidth]{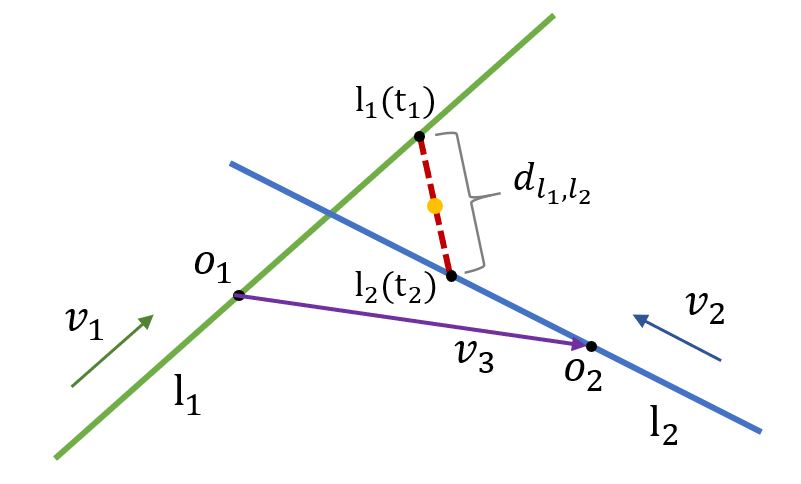}
\vspace{-8pt}
\caption{The calculation of the intersection in the 3D space. If $d_{l_1,l_2} \leqslant \epsilon$ holds, we take the midpoint of $l_1(t_1)$ and $l_2(t_2)$ as the ``intersection" point.}
\label{3d_inter}
\end{figure}

\begin{remark}
Note that this method is limited to estimate the spatial target position. When the system state $x_k$ contains other components, such as speed or acceleration, we can first determine the spatial position of the target point, then observe the final state of the agent once it reaches the target position for multiple times, and take the average as the target value of the remaining variables.
\end{remark}

\vspace{-8pt}
\section{Objective Function Identification with Inverse Optimal Control}\label{ioc}
\vspace{-8pt}
In this section, we estimate the parameters in the objective function through inverse optimal control. IOC is to identify the objective function given the optimal state or control input trajectories. In our situation, the objective function is described by $J$ in $\mathbf{P}_0$ and we are supposed to estimate the $H,R$ with $M$ optimal state trajectories in the presence of observation noises. We first formulate the inverse problem, and then analyze its well-posedness and identifiability.

\vspace{-5pt}
\subsection{Inverse Problem Formulation}
\vspace{-8pt}
According to the Pontryagin's minimum principle (PMP) for discrete-time problems \citep{bertsekas2012dynamic}, we introduce the following lemma.
\begin{lemma}\label{lem_pmp}
Consider the optimization problem $\mathbf{P}_0$.
The optimal control inputs $u_{0:N-1}^*$ and its corresponding state trajectories $x_{0:N}^*$ satisfy
\\
i) optimal control policy
$u_i^* = - R^{-1} B^T \lambda_{i+1}^*$,\\
ii) costate equation $\lambda_i^* = A^T \lambda_{i+1}^*$,\\
iii) terminal condition $\lambda_N^* = H x_N^*$,\\
with the given initial state $x_0^*$, where $\lambda_i$ is the costate of the system, $i = 0,1,\dots,N-1$.
\end{lemma}

Note that the PMP condition is a necessary condition for the optimal solution.
Therefore, in order to identify $H,R$, we set PMP condition as the constraint and formulate the following inverse problem based on $M$ trajectory observations $\{y^1,\dots,y^M\}$. Since there exist observation noises, $x_{1:N_j}$ and $\lambda_{1:N_j}$ are also optimization variables.
\begin{problem}(Inverse Optimal Control Problem)\label{prob_r}
\vspace{-8pt}
\begin{subequations}
\begin{eqnarray}\nonumber
&&\min_{\hat{H},\hat{R},x_{1:N_j}^j,\lambda_{1:N_j}^j} ~\frac{1}{M}\sum_{j=1}^M \sum_{i = 1}^{N_j} \Vert y_i^j - x^j_i \Vert^2\\ \nonumber
&&~\mathrm{s.t.} ~~x^j_{i+1} = A x^j_i - B\hat{R}^{-1}B^T \lambda^j_{i+1},\\ \nonumber
&&~~~~~~~ \lambda^j_i = A^T \lambda^j_{i+1}, \, \lambda^j_{N_j} =\hat{H} x^j_{N_j}, \, x^j_0 = y_0^j - \hat{x}_T, \\ \nonumber
&&~~~~~~~ i = 0,1,\dots,N_j-1, j = 1, \dots,M.
\end{eqnarray}
\vspace{-15pt}
\end{subequations}
\end{problem}

Note that $\mathbf{P}_0$ has the same optimal solution with the objective function $J$ replaced by $\alpha J$ for $ \alpha \in \mathbb{R}_+$. We will reveal that the inverse problem has the scalar ambiguity \citep{yu2021system} property as well. 
\begin{lemma}\label{lem_hr}
Suppose $H,R$ are the real parameters of the objective function and also the optimal solution to the Problem \ref{prob_r} with $x_{1:N}^j,\lambda_{1:N}^j$. If $H' = \alpha H, R' = \alpha R$, then $H',R',x_{1:N}^j,\alpha \lambda_{1:N}^j$ are also solutions to Problem \ref{prob_r}.
\end{lemma}
\vspace{-8pt}
\begin{pf}
The proof is given in Appendix \ref{pr1}.
\end{pf}
\vspace{-5pt}
The above analysis shows that any parameter pairs obtained through multiplying the real one $H,R$ by a non-zero scalar are all solutions to Problem \ref{prob_r}. Therefore, to simplify the problem, we set $H = I$ and estimate $R$ only during the following analysis and solution, which is reasonable in the sense that for minimizing the distance between the final state and the target point, the weights on each component of $x_N$ are always equal.

\vspace{-8pt}
\subsection{Well-posedness and Identifiability Analysis}
\vspace{-8pt}
Before estimating the parameters, we first prove the well-posedness of the problem, i.e., whether the optimal $R$ is unique. Denote $A^c_k = A - B K_k$ as the close-loop system matrix.

\begin{thm}\label{thm_wp}
For problem $\mathbf{P}_0$, if there exist two positive definite matrices $R,R'$ generate the same closed-loop control system matrix $A^c_k$, then we have $R = R'$.
\end{thm}
\vspace{-8pt}
\begin{pf}
The proof is given in Appendix \ref{pr2}.
\end{pf}
\vspace{-5pt}

We have showed the well-posedness of the inverse problem. Now we need to prove that we can obtain the true value of $R$ by solving Problem \ref{prob_r} when the amount of observations is large enough.
\begin{thm}\label{thm_r}
Suppose $\hat{R}, x_{1:N}^*,\lambda_{1:N}^*$ are the optimal solution to Problem \ref{prob_r}. As $M \xrightarrow{} \infty$, we have $\hat{R} \xrightarrow{\mathrm{P}} R$, where $R$ is the true parameter in the forward problem.
\end{thm}
\vspace{-8pt}
\begin{pf}
The proof is given in Appendix \ref{pr3}.
\end{pf}

\vspace{-8pt}
\section{Control Horizon Estimation and Control Input Prediction}\label{horizon}
\vspace{-8pt}
\subsection{Control Horizon Estimation}
\begin{figure}[t]
\centering
\includegraphics[width=0.3\textwidth]{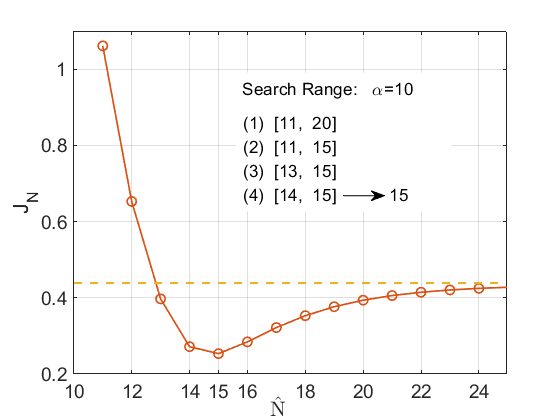}
\vspace{-8pt}
\caption{Set the observation number $l = 10$ and the real control horizon $N=15$. The red curve shows the change of function $J_N$ with $\hat{N}$. $J_N$ decreases at the beginning and reaches the minimum value at $15$, then gradually increases and converges to $\sum_{i = 1}^{10} \Vert y_i\Vert^2$.}
\label{n_est}
\end{figure}

\vspace{-8pt}
After obtaining the target state $\hat{x}_T$ and the parameter $\hat{R}$, we need to estimate the control horizon $N$. We provide the following optimization problem for the estimation.
\begin{problem}(Estimation of the Control Horizon)\label{pro_n}
\vspace{-12pt}
\begin{subequations}
\begin{eqnarray}\nonumber
&&\min_{\hat{N}} ~~\sum_{k = 1}^l \Vert y_k - x_i \Vert^2 : = J_N(\hat{N};y_{0:l})\\ \nonumber
&&~\mathrm{s.t.} ~~ \eqref{sys},u_i = -K_i x_i,\,x_0 = y_0- \hat{x}_T,\\ \nonumber
&&~~~~~~~ K_i =  (\hat{R} + B^T P_{i+1} B)^{-1} B^T P_{i+1} A, \\ \nonumber
&&~~~~~~~ P_i =K_i^T \hat{R} K_i + {A^c_i}^T P_{i+1} A^c_i, \,P_{\hat{N}} = I,\\\nonumber
&&~~~~~~~
i = 0,1,\dots,\hat{N}-1,
\end{eqnarray}
\end{subequations}
where $y_{1:l}$ is the observation of $R_M$ up to time $k=l$.
\end{problem}
The above problem reflects that the trajectory deviation from the observed data $y_{1:l}$ is the smallest under the optimal solution $\hat{N}^*$. Since the optimization variable $\hat{N}$ does not explicitly exist in the problem and since $\hat{N} \in \mathbb{N}_+$, Problem 2 is a non-convex optimization problem over the positive integers and hard to solve directly. Therefore, we first investigate how the value of the objective function changes with $\hat{N}$. Note that $N > l$ and as $\hat{N}$ grows from $l$ to $N$, the objective function $J_N$ gradually decreases to the minimum value. Then after the optimal value, $J_N$ increases with the growth of $\hat{N}$, and finally converges to a fixed value $\sum_{i = 1}^l \Vert y_i\Vert^2$. See an illustration in Section \ref{sim} Fig. \ref{n_est}.

According to the analysis of $J_N$, to obtain the solution of Problem \ref{pro_n}, we can traverse from $\hat{N} = l$ and keep increasing $\hat{N}$ until $J_N$ no longer decreases, at which point the optimal $\hat{N}^*$ is found. However, this requires lots of computation if $N \gg l$. Therefore, we propose an algorithm based on binary search to find the optimal solution inspired by the line search of the gradient decent method. Since $J_N$ is a discrete function with respect to $\hat{N}$, we use the function values at both $\hat{N}$ and $\hat{N}+1$ to approximate the gradient at point $\hat{N}$, which is given by:
\begin{equation}\label{gradient}
g_N = J_N(\hat{N}+1;y_{0:l})-J_N(\hat{N};y_{0:l}).
\end{equation}
Thus, if we have $g_N < 0$, then $\hat{N} < \hat{N}^*$; if $g_N>0$, then $\hat{N} \geqslant \hat{N}^*$.
The detailed algorithm is shown in Algorithm \ref{alg1}.

\begin{algorithm2e}\label{alg1}
\SetAlgoLined
 \caption{Binary Search for Optimal $\hat{N}$}
    \KwIn{ 
    The observation trajectory, $y_{0:l}, l$;
      The estimate of the target state, $\hat{x}_T$;
      The system parameters, $A,B$;
      The estimate of the objective function, $\hat{R}$; The step length, $\alpha$;}
    \KwOut{
      The optimal control horizon estimate, $\hat{N}^*$;}
    \textbf{Determine the initial bound:}\\
    Set the lower bound as $N^{-} = l+1$;
    Let $\hat{N}' = l + \alpha$;\\
    \lWhile{$g_{\hat{N}'} < 0$}{
    $\hat{N}' = \hat{N}' + \alpha$}
    Set the upper bound as $N^+ = \hat{N}'$;\\
    \textbf{Binary search:}\\
    Take the midpoint of the range as $\Tilde{N} = \lfloor{\frac{N^- + N^+}{2}}\rfloor$;\\
    \While{$N^+ - N^- > 1$}{
    \lIf{$g_{\Tilde{N}} > 0$} {Set $N^+ = \Tilde{N}$}
    \lElse{Set $N^- = \Tilde{N}$}
    }
    $\hat{N}^* = \mathop{\arg\min}\limits_{\hat{N}} \{J_N(\hat{N};y_{0:l}); \hat{N} \in \{N^+,N^-\} \}$;\\
\textbf{return} The control horizon estimation $\hat{N}^*$.
\end{algorithm2e} 

\begin{remark}
Notice that after determining the initial range $[N^-,N^+]$, if we simply leverage the function values of the two break points $N_1,N_2,N^-\leqslant N_1<N_2 \leqslant N^+$ to find the optimum, to have a constant compressive ratio $c$, the break points should satisfy $\frac{N_2-N^-}{N^+ - N^-}= \frac{N^+ - N_1}{N^+ - N^-} = c$ and $c = \frac{\sqrt{5}-1}{2} \approx 0.618$. However, with Algorithm \ref{alg1}, since we approximate the gradient, the compressive ratio is improved to $c = 0.5$. Moreover, in order to reduce the computation cost, it is recommended to store the result each time we calculate the deviation sum $J_N$ corresponding to a certain $\hat{N}$ to avoid repeated calculation.
\end{remark}
\vspace{-8pt}
\subsection{Optimization Problem Reconstruction}
\vspace{-8pt}
Now we have obtained the estimates of the target state $\hat{x}_T$, the control horizon $\hat{N}^*$, and identified the objective function $J$. 
Therefore, we can infer the control input of the target agent $R_M$ by 
\begin{equation}
\hat{{u}}_{l|l} = \mu_0,
\end{equation}
where $\mu_0$ is calculated by solving the following problem:
\begin{problem}(Control Input Inference)\label{p1}
\vspace{-12pt}
\begin{subequations}
\begin{eqnarray}\nonumber
&&\min_{\mu_{0:N'-1}} J' = x_{N'}^T  x_{N'} + \sum_{k = 0}^{N'-1} \mu_k^T \hat{R} \mu_k\\ &&\mathrm{s.t.} ~~\eqref{sys}, x_0 = \hat{x}_{l|l} - \hat{x}_T, k = 0,1,\dots,N'-1,\nonumber
\end{eqnarray}
\end{subequations}
\end{problem}
in which $N' = \hat{N}^*-l$.
The solution is given in \cite{kwakernaak1972maximally}:
\begin{equation}\label{mu0}
\mu_0 = - K_0 x_0,
\end{equation}
where for $k = 0,1,\dots,N'-1$,
\begin{equation}\label{pk_iter}
\left\{
\begin{array}{ll}
K_k \!=\!  (\hat{R} + B^T P_{k+1} B)^{-1} B^T P_{k+1} A,,\\
P_k \!=\!K_k^T\hat{R} K_k \!+\! {A^c_k}^T P_{k+1} {A^c_k}, \, P_{N'} = I.
\end{array}
\right.
\end{equation}
\begin{remark}
According to the principle of optimality \citep{bellman1966dynamic}, if a control policy $p_{0,N}^*$ is optimal for the initial point $x_0$, then for any $l\in \{1,2,\dots,N-1\}$, its sub-policy $p_{l,N}^*$ is also optimal to the subprocess containing last $N-l+1$ steps with the initial point $x_l$.
\end{remark}


\begin{algorithm2e}\label{alg2}
 \caption{Control Inputs Inference Algorithm}
    \KwIn{ The observation data including $M$ history trajectories, $\{y^{1:M}\}$;
    The observation of current trajectory and the observation times, $y_{0:l}, l$;
    The system dynamic function, $A,B$;}
    \KwOut{
      The control input inference at time $l$, $\hat{u}_{l|l}$;}
    Estimate the target state $\hat{x}_T$ through line fitting and calculating the intersection;\\
    Identify the objective function parameter $\hat{R}$ by solving Problem \ref{prob_r} with $M$ trajectories ;\\
    Calculate the optimal estimate to control horizon $\hat{N}^*$ with Algorithm \ref{alg1};\\
    Formulate and solve Problem \ref{p1} by dynamic programming with the previous estimates $\hat{x}_T$, $\hat{R}$, $\hat{N}^*$; Obtain the one-step input $\mu_0$ with \eqref{mu0} in the forward pass;\\
    \textbf{return} The control input inference $\hat{u}_{l|l} = \mu_0$.
\end{algorithm2e} 
\vspace{-8pt}
\section{Simulation Results}\label{sim}
\vspace{-8pt}
We conduct multiple simulations on our algorithm to show the performance and efficiency. Consider a mobile agent modeled by the following dynamical system:
\begin{equation}\label{sim_sys}
x_{k+1} = A x_k + B u_k = \begin{pmatrix}
    1.2 &0\\
    0& 1.2
\end{pmatrix} x_k + \begin{pmatrix}
    1 &0\\
    0& 1
\end{pmatrix}u_k,
\end{equation}
where $n = m =2$.
Assume the agent is driving from $(0,0)^T$ to $(6,8)^T$. The control optimization problem is the same as $\mathbf{P}_0$ and we have
$H = I, R = \begin{pmatrix}
    0.5 &0\\
    0& 0.5
\end{pmatrix}$.
Through applying the external incentives and the line fitting, we calculate the intersection point as the estimation to the target state $\hat{x}_T = (6.063,8.086)^T$.

\begin{figure}[t]
\centering
\includegraphics[width=0.3\textwidth]{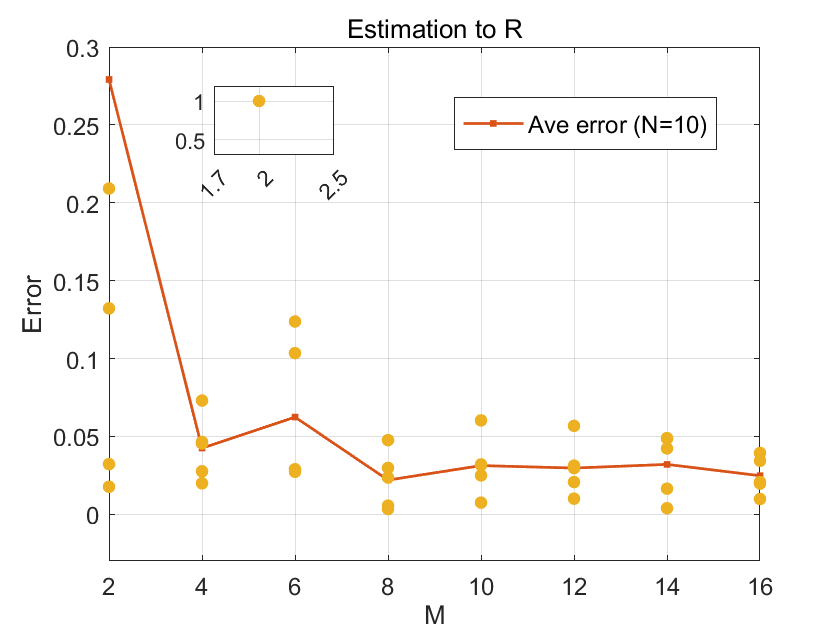}
\vspace{-8pt}
\caption{Estimation results of the parameter $R$. We perform the simulation $5$ times at each $M$. The yellow dots are estimation errors of each simulation and the red curve represents the change of the average error from $M=2$ to $16$.}
\label{q_est}
\end{figure}

To test the estimation algorithm for $R$, we set $N_j=10$ and give random initial states $x_0^j$ for all $j = 1,2,\dots,M$. The observation noises satisfy a Gaussian distribution $\mathcal{N}(0,0.02^2)$.  Use the MATLAB function $fmincon$ to solve Problem \ref{prob_r}. The estimation error is calculated as $\frac{\Vert \hat{R}-R \Vert}{\Vert R \Vert}$. The results are shown in Fig. \ref{q_est} and the time costs are listed in Table. \ref{tab1}. Since $R$ is a two-dimensional matrix, more than two trajectories are required to solve for a unique $\hat{R}$. We can see that as the number of trajectories $M$ increases from $2$ to $16$, the estimation error shows a decreasing trend and stabilizes at a small value when $M \geqslant 8$. Therefore, considering both of the estimation error and the computational efficiency, we choose $M=4$ and the estimation value $\hat{R} = \begin{pmatrix}
    0.5772 & -0.0521\\
    -0.0521& 0.5242
\end{pmatrix}$.

\begin{table}
    \centering
    \vspace{-8pt}
    \caption{Time cost of $R$ estimation}
	\label{tab1}
    \begin{tabular}{cccccc}
    \toprule    
    $M=$ & 2 & 4 & 6 & 8 & 10 \\    
    \midrule   
    Time Cost (s) & 0.5436 & 0.7829 & 1.071 & 1.431 & 1.806\\
    \bottomrule   
    \end{tabular}
\end{table}

Now we start to infer the inputs and predict the future states. Observe for $l =10$ and set $\alpha = 10$. With Algorithm \ref{alg1}, we obtain the optimal control horizon $\hat{N}^* = 15$. The current state is estimated by Kalman filter as $\hat{x}_{l|l}$. Then we reconstruct and solve the control optimization problem of the mobile agent as Problem \ref{p1}:
\begin{subequations}
\vspace{-10pt}
\begin{eqnarray}\nonumber
&&\min_{\mu_{0:4}} x_{5}^T \begin{pmatrix}
    1 & 0\\
    0& 1
\end{pmatrix}x_{5} + \sum_{k = 0}^{4} \mu_k^T \begin{pmatrix}
    0.5772 & -0.0521\\
    -0.0521& 0.5242
\end{pmatrix} \mu_k\\ &&\mathrm{s.t.} ~~\eqref{sim_sys}, x_0 = \hat{x}_{l|l} - (6.063,8.086)^T, k = 0,1,\dots,4.\nonumber
\end{eqnarray}
\end{subequations}
Note that the solution $x_{1:5}$ of above problem is actually the prediction of the future state of the agent from $k=11$ to $15$. Denote the prediction error as $\Vert \hat{x} - x\Vert$. The comparisons with the prediction through polynomial regression \citep{chen2016tracking} with same observation noise distribution are shown in Table. \ref{tab2}. The curve fitting is based on all $l=10$ history states and the optimal highest polynomial order is $3$. We can see the prediction error of our methods is overall lower than the fitting methods.

\begin{table}
    \centering
    \caption{States prediction error comparison}
	\label{tab2}
 \resizebox{\columnwidth}{!}{
    \begin{tabular}{cccccc}
    \toprule    
    $k=$ & 11 & 12 & 13 & 14 & 15 \\    
    \midrule   
    \textbf{Ours} & 0.0244 & 0.0180 & 0.0123 & 0.0071 & 0.0032\\
    \cite{chen2016tracking} & 0.0362 & 0.0738 & 0.1337 & 0.2252 & 0.3606\\
    \bottomrule   
    \end{tabular}}
\end{table}
\vspace{-8pt}
\section{Conclusion And Future Work}\label{conclusion}\label{conc}
\vspace{-8pt}
This paper proposes an algorithm for the control input inference of a mobile agent without knowing its objectives. Assuming the linear dynamical system is known and the agent has a quadratic objective function with unknown parameters, we first estimate the target state. Then we identify the parameters in the objective function based on inverse optimal control and estimate the control horizon with binary search. Finally, we reconstruct the control optimization problem of the agent and calculate its solution as the prediction to the input. 

\appendix
\vspace{-8pt}
\section{Proof of Lemma \ref{lem_hr}}\label{pr1}
\vspace{-8pt}
Applying the problem constraints to the new parameters $H',R'$, we obtain that $(\lambda_i^j)' = \alpha \lambda_i^j$ for all $i,j$ and
\begin{align}\nonumber
(x^j_{i+1})'& = A (x^j_i)' - B(R')^{-1}B^T (\lambda^j_{i+1})'\\ \nonumber & = A (x^j_i)' - BR^{-1} B^T \lambda_{i+1}^j.
\end{align}
Comparing to $x^j_{i+1} = A x^j_i - BR^{-1} B^T \lambda_{i+1}^j$, if we have $(x_0^j)' = x_0^j$, it is easy to realize that $H',R',x_{1:N}^j,(\lambda_{1:N}^j)'$ are also solutions to the problem with the same objective function value of the real parameters $H,R$.
\vspace{-8pt}
\section{Proof of Theorem \ref{thm_wp}}\label{pr2}
\vspace{-8pt}
Solving $\mathbf{P}_0$ with dynamic programming, we have \eqref{eq:uk} and
\begin{align}\label{K}
& K_k =  (R + B^T P_{k+1} B)^{-1} B^T P_{k+1} A,\\
& P_k =K_k^TR K_k + {A^c_k}^T P_{k+1} A^c_k, \,P_{N} = H.\label{P}
\end{align}
Suppose there exist two different positive definite matrices $R,R'$ that can generate the same close-loop system and we have $R' = R + \Delta R$, where $\Delta R$ is a symmetric matrix. Then there are $P_{0:N}, P_{0:N}'$ and $K_{0:N-1}, K_{0:N-1}'$ satisfying \eqref{K} and \eqref{P}. Since their close-loop system matrices are the same, which means $A^c_k = {A^c_k}'$ for all $k$, we have
\[
A-B K_k \! =\! A-BK_k' \Leftrightarrow B K_k \!=\! BK_k' \Leftrightarrow B^TB K_k \!=\! B^T B K_k'.
\]
Note that $B$ has full column rank, thus $B^TB$ is invertible. Therefore, we have $K_k = K_k'$.

According to \eqref{K}, it follows that
\begin{align}\nonumber
(R + B^T P_{k+1} B) K_k = B^T P_{k+1} A\Rightarrow 
R K_k = B^T P_{k+1} A^c_k.
\end{align}
Then we also have the above equation for $R'$ and $P_{k+1}'$:
\begin{align}\nonumber
(R + \Delta R) K_k& = B^T (P_{k+1} + \Delta P_{k+1}) A^c_k,\\
\Delta R K_k& = B^T \Delta P_{k+1} A^c_k. \label{delta_r}
\end{align}
Similarly, for \eqref{P} we have
\begin{align}\nonumber
P_k + \Delta P_k& =K_k^T (R + \Delta R) K_k + {A^c_k}^T (P_{k+1} +\Delta P_{k+1} A^c_k),\\
\Delta P_k& = (K_k^T B^T + {A^c_k}^T) \Delta P_{k+1} A^c_k.\label{delta_p}
\end{align}
Since $P_N = P_N' = H$, then $\Delta P_N = 0$. Combining with \eqref{delta_p}, we have $\Delta P_k = 0$ and $P_k = P_k'$ for all $k$. Thus \eqref{delta_r} converts into
$
\Delta R K_k = 0.
$
Note that $H\succ 0, R\succ 0$, then with \eqref{P} we can obtain that $P_k \succ 0$ and is invertible. Therefore, since $P_k$ and $A$ are all invertible matrices, from \eqref{K} we have $rank(K_k) = rank(B^T) = m$. Thus $K_k$ has full row rank and $\Delta R =0$.
\vspace{-8pt}
\section{Proof of Theorem \ref{thm_r}}\label{pr3}
\vspace{-8pt}
The proof idea is similar to Theorem 4.1 in \cite{zhang2019inverse}, while the specific problems are different. Therefore, we will show how to transform our problem to theirs and then the proof immediately follows.

Denote $z_i = \begin{pmatrix} x_i^T & \lambda_i^T
\end{pmatrix}^T, i = 1, \dots,N$. Then the constraints for each step in Problem \ref{prob_r} is written as
\begin{equation}
\underbrace{\begin{bmatrix}
I & B R^{-1} B^T \\ 0 & A^T
\end{bmatrix}}_E z_{i+1} = 
\underbrace{\begin{bmatrix}
A & 0 \\ 0 & I
\end{bmatrix}}_F z_i.
\end{equation}
Combine all the constraints into a matrix equation as
\begin{equation}\label{fzb}
\vspace{-5pt}
\underbrace{\begin{bmatrix}
\Tilde{E} & & & \Tilde{F} \\
-F & E & & \\
 & \ddots & \ddots & \\
 & & -F & E \\
\end{bmatrix}}_{\mathscr{F}(R)}
\underbrace{\begin{bmatrix}
z_1 \\ z_2 \\ \vdots \\ z_N
\end{bmatrix}}_Z
= \underbrace{\begin{bmatrix}
A \\ 0 \\ \vdots \\ 0
\end{bmatrix}}_{\Tilde{A}}x_0,
\end{equation}
where $\Tilde{E} = \begin{bmatrix}
I & B R^{-1} B^T \\ 0 & 0
\end{bmatrix}$ and $\Tilde{F} = \begin{bmatrix}
0 & 0 \\ -I & I
\end{bmatrix}$. Note that $\mathscr{F}(R)$ is an invertible matrix. Then we have
\begin{equation}\nonumber
\vspace{-5pt}
\sum_{i = 1}^N \Vert y_i - x^*_i \Vert^2 = \Vert Y - G_X Z \Vert^2 = \Vert Y - G_X \mathscr{F}(R)^{-1} \Tilde{A} x_0 \Vert^2,
\end{equation}
where $G_X = I_{N-1} \otimes [I_n, 0_n]$. 
Then the subsequent proof is the same as \cite{zhang2019inverse} after replacing $\mathscr{F}(Q)$ with $\mathscr{F}(R)$.

\vspace{-8pt}
\bibliography{ref}
\vspace{-10pt}
\end{document}